\documentclass[times,twocolumn,final]{elsarticle}
\usepackage{medima}
\usepackage{framed,multirow}
\usepackage[numbers]{natbib}
\usepackage{subfig}
\usepackage{amsmath,amssymb,amsfonts,amsthm}
\usepackage{algorithmic}
\usepackage{graphicx}
\usepackage{hyperref}
\usepackage{multirow}
\usepackage{multicol}
\usepackage{booktabs}
\usepackage{makecell}
\usepackage{tabularx}
\usepackage{gensymb}
\hypersetup{colorlinks=true,urlcolor=red}
\usepackage{bm} 
\usepackage{xspace}

\newcommand{\beq}{\begin{equation}}
\newcommand{\eeq}{\end{equation}}
\newcommand{\beqa}{\begin{eqnarray}}
\newcommand{\eeqa}{\end{eqnarray}}

\newcommand{\add}[1] {\textcolor{black}{#1}} 

\usepackage{pifont}
\newcommand{\cmark}{\ding{51}}
\newcommand{\xmark}{\ding{55}}
\newcommand{\mbtx}{{\mathbf{\tilde x}^{\text{emb}}_{\text{txt}}}}
\newcommand{\mbty}{{\mathbf{\tilde y}^{\text{emb}}_{\text{txt}}}}
\newcommand{\mbey}{{\mathbf{y}^{\text{emb}}_{\text{txt}}}}
\newcommand{\mbx}{{\mathbf{x}^{\text{emb}}_{\text{txt}}}}
\newcommand{\mby}{{\mathbf{y}_{\text{txt}}}}

\newcommand{\mbix}{{\mathbf{x}_{\text{img}}}}
\newcommand{\mbiy}{{\mathbf{y}_{\text{img}}}}

\newcommand{\correct}[1] {\textcolor{blue}{#1}}

\usepackage{marginnote}

\journal{Medical Image Analysis}

\begin{document}
\verso{K. Kim, Y. Oh, and S. Park \textit{et~al.}}
\begin{frontmatter}
\title{End-to-End Breast Cancer Radiotherapy Planning via  LMMs with Consistency Embedding}

\author[1]{Kwanyoung \snm{Kim}\corref{cor1}} 
\author[2]{Yujin \snm{Oh}\corref{cor1}}
\author[3,7]{Sangjoon \snm{Park}\corref{cor1}}
\author[4]{Hwa Kyung \snm{Byun}}
\author[5]{Joongyo \snm{Lee}}
\author[3,6]{Jin Sung \snm{Kim}\corref{cor2}}
\author[3]{Yong Bae \snm{Kim}\corref{cor2}}
\author[8]{Jong Chul \snm{Ye}\corref{cor2}}

\address[1]{Samsung Research, Seoul 06765, Republic of Korea}
\address[2]{Center for Advanced Medical Computing and Analysis (CAMCA), Department of Radiology, Massachusetts General Hospital (MGH) and Harvard Medical School, MA 02114, USA}
\address[3]{Department of Radiation Oncology, Yonsei University College of Medicine, Yonsei University, Seoul 03772, Republic of Korea}
\address[7]{Institute for Innovation in Digital Healthcare, Yonsei University, Seoul 03772, Republic of Korea}
\address[4]{Department of Radiation Oncology, Yongin Severance Hospital, Yongin-si, Gyeonggi-do 16995, Republic of Korea}
\address[5]{Department of Radiation Oncology, Gachon University Gil Hospital, Incheon 21565, Republic of Korea}
\address[6]{Oncosoft Inc, Seoul 03776, Republic of Korea}
\address[8]{Kim Jaechul Graduate School of AI, Korea Advanced Institute of Science and Technology (KAIST), Daejeon 341431, Republic of Korea}

\ead{jong.ye@kaist.ac.kr}
\cortext[cor1]{Co-first authors}
\cortext[cor2]{Co-Corresponding authors}

\begin{abstract}
\marginparsep 20pt
Recent advances in AI foundation models have significant potential for lightening the clinical workload by mimicking the comprehensive and multi-faceted approaches used by medical professionals. In the field of radiation oncology,  the integration of multiple modalities holds great importance, so the opportunity 
of foundational model is abundant. Inspired by this, here we present RO-LMM, a multi-purpose, comprehensive large multimodal model (LMM) tailored for the field of radiation oncology. This model effectively manages a series of tasks within the clinical workflow, including clinical context summarization, \add{radiotherapy strategy} suggestion, and plan-guided target volume segmentation by leveraging the capabilities of LMM. In particular, to perform consecutive clinical tasks without error accumulation, we  present a novel Consistency Embedding Fine-Tuning (CEFTune) technique, which boosts LMM's robustness to noisy inputs while preserving the consistency of handling clean inputs. We further extend this concept to LMM-driven segmentation framework, leading to  a novel  Consistency Embedding Segmentation~(CESEG) techniques. Experimental results including multi-centre validation confirm that our RO-LMM with CEFTune and CESEG results in promising performance for multiple clinical tasks with generalization capabilities. 
\end{abstract}


\end{frontmatter}

\begin{figure*}[htb!]
\centering
    \captionsetup{type=figure}
    \includegraphics[width=0.9\linewidth]{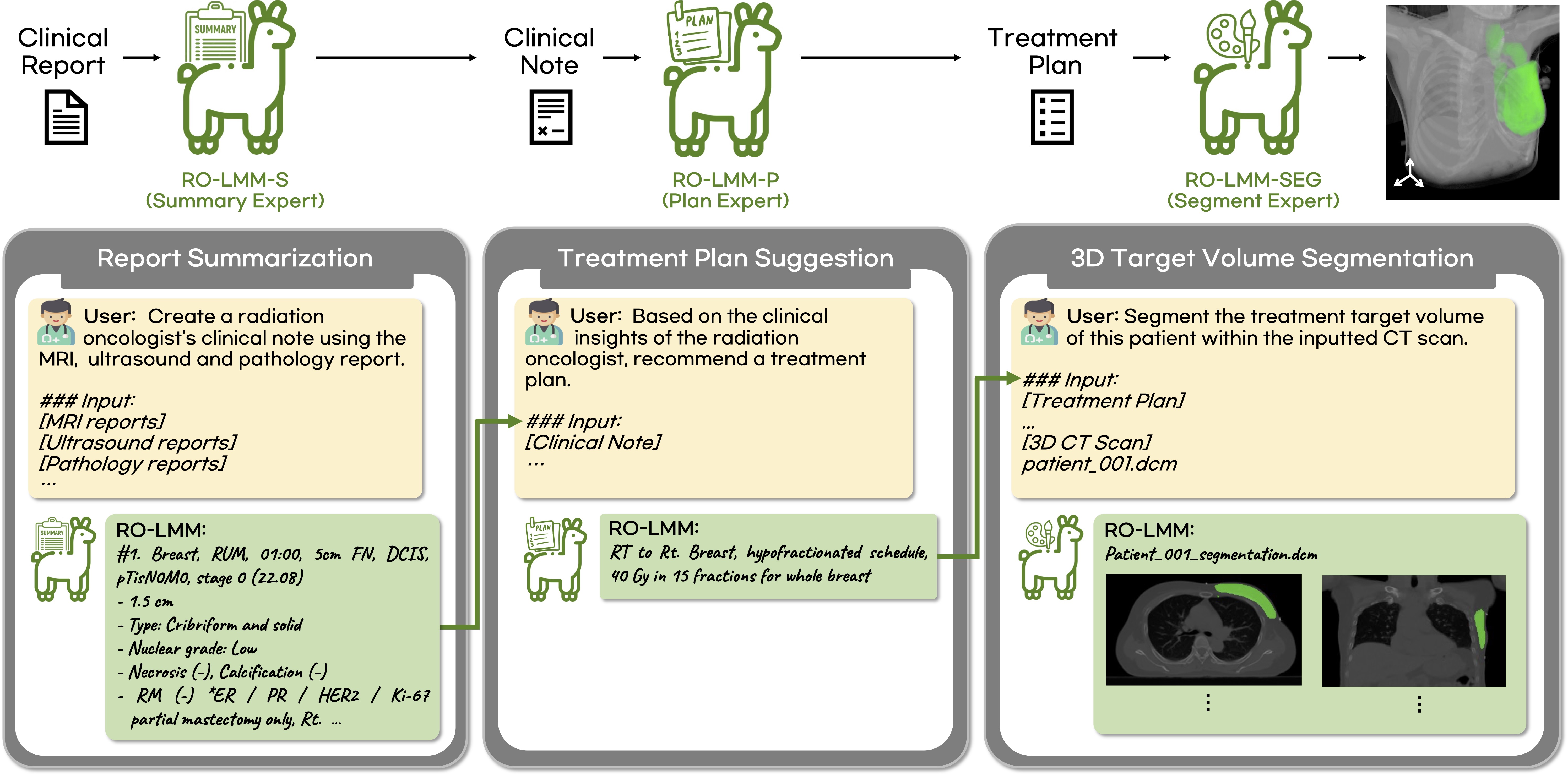}
    \captionof{figure}{RO-LMM as an assistant large multimodal model (LMM) in the field of radiation oncology. The model seamlessly covers  various tasks such as clinical report summarization, radiation \add{radiotherapy strategy} suggestion, and 3D target volume segmentation.} 
    \label{intro}
\end{figure*}

\section{Introduction}

\label{sec:intro}

Recently, the emergence of a new generation of AI models inspired by large language models (LLM), known as foundation models, marks a significant departure from previous paradigms \cite{moor2023foundation}. 
These models are characterized by their massive scale and versatility, which stems from self-supervised training on a diverse array of data. 
These foundation models are now capable of achieving state-of-the-art performance (SOTA) in a wide range of domains, including tasks such as multimodal reasoning, image-to-text generation, image captioning, and text-guided image segmentation \cite{bubeck2023sparks, dai2024instructblip, driess2023palme, li2023blip, liu2024visual, lai2024lisa}.

These characteristics signify a potential paradigm shift in how AI can be integrated into medical practices, which inherently rely on multimodal information for comprehensive clinical decision making. 
Furthermore,  this gives an opportunity of overcoming limitations of  now over 500 FDA-approved AI models, which are mainly specialized for a specific task with unimodal information~\cite{joshi2024fda}. 
Specifically, in contrast to these unimodal AIs, 
 generalist medical AIs that synergistically combine foundation models can encompass a holistic understanding of clinical workflows, which can receive  a variety of medical data, including imaging modalities,  \add{electronic health records, laboratory results, genomics, and even clinical reports \cite{singhal2023large, rajpurkar2023current, wu2023towards,moor2023foundation,tu2024towards}.}
 By understanding various types of data and their interrelationships, multimodal AIs would facilitate more accurate diagnoses, personalized treatment development, and a reduction in medical errors by providing a comprehensive view of patient data. 
 
\add{In radiation oncology, the primary focus of this article, the integration of multiple modalities is crucial, making it one of the most important clinical fields for evaluating the potential of foundation models.}
Accordingly, here we introduce RO-LMM, a prototype large multimodal model (LMM) specifically designed to support the clinical workflow in radiation oncology. In particular, this work significantly extends our prior related work called LLMSeg\cite{oh2023llm}, which focused on multimodal segmentation. 
More specifically, RO-LMM enlarges the LLMSeg's scope by tackling a broader range of clinical tasks in radiation oncology: (1) It efficiently summarizes extensive patient histories and examination results into concise but informative clinical notes. Additionally, it is capable of (2) proposing appropriate \add{radiotherapy strategies} from a clinical expert perspective and (3) delineating radiation target volumes on 3-dimensional (3D) computed tomography (CT) scans consistent with the proposed \add{radiotherapy strategies}. \add{This multifaceted functionality of RO-LMM demonstrates significant advances in supporting the expertise of clinical professionals.}

In training LLM to perform these sequential tasks from \add{radiotherapy strategy} suggestion to target volume segmentation, {we found that there exist  potentials for error accumulation from each task which can potential leads to significant end-to-end performance loss.} Accordingly, another important contribution of this work is
the employment and extension of a Noisy Embedding Fine-Tuning (NEFTune) \cite{jain2024neftune}  which involves injecting uniform noise into embeddings during the training for  each targeted task. More specifically, to further enhance the model's applicability, we develop a novel  Consistency Embedding Fine-Tuning (CEFTune) technique, which add regularization loss to enforce the consistency between the prediction given noisy and clean inputs. Furthermore, by expanding beyond text-related tasks, we apply these concepts to 3D segmentation tasks, resulting in novel Noisy Embedding Segmentation (NESEG) and Consistency Embedding Segmentation (CESEG).  \add{These advancements prevent error propagation between subsequent tasks and collectively contribute to a marked improvement in the end-to-end model's generalization capabilities in both internal and external validation.}

As for the proof of the concept study,  our RO-LMM framework was
adopted for breast cancer study,   one of the high-prevalence cancer types requiring relatively standardized radiotherapy based solely on CT imaging. 
Our contributions can be summarized as:
\begin{itemize}
\item We propose a comprehensive framework, dubbed as RO-LMM, wherein LMM assists the broad workflow of radiation oncology for breast cancer treatment. To the best of our knowledge, our prototype is the first to assist in the comprehensive workflow of radiation oncology.
\item To prevent from potential error accumulation throughout consecutive clinical tasks such as clinical context summarization, \add{radiotherapy strategy} suggestion, and plan-guided target volume segmentation, we explore noise augmentation and consistency, and propose a novel training approach, such as CEFTune, NESEG, and CESEG that significantly enhance the robustness of our method.
\item Through experiments on various validation settings with real clinical data of breast cancer patients, we demonstrated that RO-LMM outperforms Default methods.
\end{itemize}

\begin{figure*}[ht!]
\centering
\includegraphics[width=0.9\linewidth]{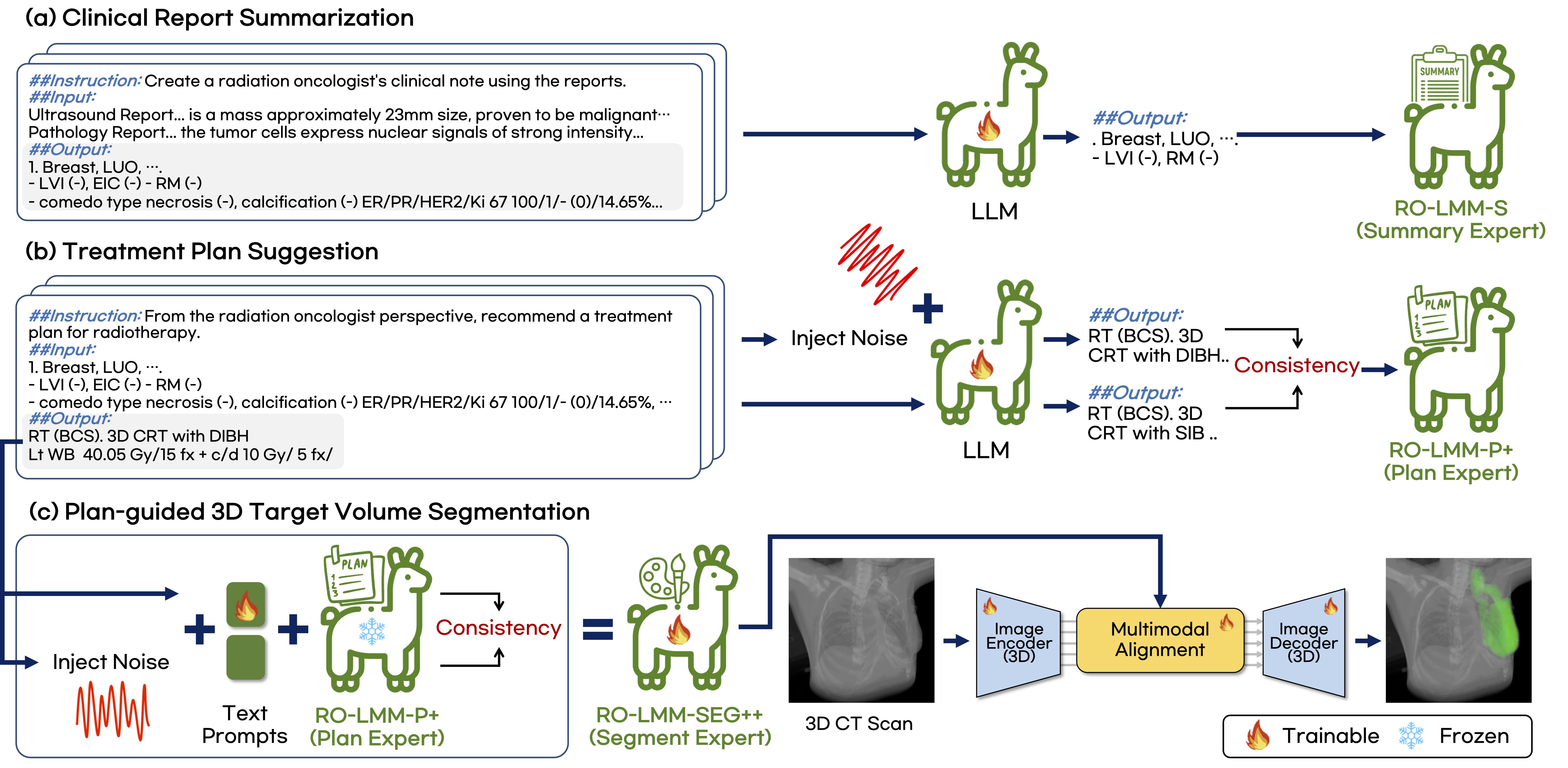}
\caption{Schematics of RO-LMM training for three different tasks. (a) RO-LMM-S for clinical note summarization. (b) RO-LMM-P++ for \add{radiotherapy strategy} suggestion. (c) RO-LMM-SEG++ for plan-guided target volume segmentation.}
\label{fig_schematic1}
\end{figure*}

\section{Related Works}

\subsection{Instruction Fine-tuning in LLM} 
In the field of LLM, the instruction fine-tuning has been emerged as a pivotal technique for augmenting model responsiveness. The simplicity and effectiveness of this approach have been introduced in numerous works, with a notable scale-up facilitated by advancements of large language models (LLMs) such like GPT-3~\cite{brown2020language}, ChatGPT~\cite{chatgpt}, and GPT-4~\cite{OpenAI2023GPT4TR}. Notably, a pioneering contribution, Self-Instruct~\cite{wang2022self}, involves fine-tuning of foundation models using instruction-output pairs generated from InstructGPT~\cite{ouyang2022training}. This methodology, along with similar approaches, has led to diverse language model variants, including Alpaca~\cite{taori2023stanford}, Vicuna~\cite{chiang2023vicuna}, Dolly~\cite{DatabricksBlog2023DollyV2}, and LLaMA2~\cite{touvron2023llama2}, by exhibiting promising performance across a wide range of tasks.
In the medical domain, Chat-Doctor~\cite{li2023chatdoctor}, Med-Alpaca~\cite{han2023medalpaca}, PMC-LLaMA~\cite{wu2023pmc}, and Asclepius~\cite{kweon2023publicly} have been fine-tuned for clinical question-answering (QA) task. Although these models demonstrate robust performance for diverse QA benchmarks, there is a notable gap in their applicability for the practical clinical workflow in a specific domain, such as radiation oncology, which serves as our targeted focus.

\subsection{Stabilizing LLM Fine-tuning with Noise} 
To improve the robustness of language models against noisy input, various strategies have been explored. Approaches such as SMART~\cite{jiang2019smart}, FreeLB~\cite{zhu2020FreeLB}, and R3F~\cite{aghajanyan2020better}, LSNR~\cite{hua2021noise} utilize adversarial training by introducing minor Gaussian perturbations in embedding dimensions, optimizing model performance in noisy environments. 
More recently, NEFTune~\cite{jain2024neftune}  improves LLM performance during fine-tuning by simply adding random noise into the embedding vectors during training. In the context of this study, we extend NEFTune by incorporating the concept of consistency regularization. This extension aims to further improve the robustness and generalization capabilities of language models when faced with noisy input.

\subsection{Language-driven Image Segmentation} 

Recent research in the field of image segmentation has emerged to incorporate linguistic ability, such like language-driven semantic segmentation \cite{li2022languagedriven}, open-vocabulary segmentation \cite{ding2023open,  liang2023open}, referring segmentation \cite{wang2022cris}, and reasoning segmentation \cite{lai2024lisa}.
Language-driven segmentation has triggered a paradigm shift in medical domain, where multimodal knowledge is inevitable. For instance, LViT~\cite{10172039} and ConTEXTualNet~\cite{huemann2023contextual} introduce text-driven chest X-ray radiography segmentation. 
Target volume segmentation in the field of radiation oncology is more challenging, due to its intrinsic need for considerations of the clinical aspects beyond the image, such as overall cancer stage, treatment aim, pathological findings, and so on \cite{choi2020clinical, chung2021clinical, hosny2018artificial}. Recent research on clinical context-aware breast cancer radiotherapy delineation \cite{oh2023llm} has demonstrated that the multimodal AI outperforms the traditional unimodal AI by a substantial margin, particularly when labeled datasets are scarce. By extending \cite{oh2023llm}, we incorporate \add{radiotherapy strategies} generated by LMM directly from clinical reports, which aligns seamlessly with actual clinical workflows in radiation oncology.

\section{Methods}

In this section, we provide a detailed description of our proposed approach designed for sequential text generation tasks, including summarization and suggestions, as well as text-driven image segmentation, whose robustness is improved by consistency embedding finetuning.  The overall framework is illustrated in Figure~\ref{fig_schematic1}.

\subsection{Consistency Embedding Fine-tuning for Clinical LMM} \label{sec:text}

To realize the multi-purpose LMM with expertise in clinical report summarization and \add{radiotherapy strategy} suggestion, we conduct instruction fine-tuning for LLaMA2~\cite{touvron2023llama2}.
Considering the nuanced differences in 
the intended objective of each task, we adopt separate training strategies to acquire task-specific expertise, namely RO-LMM-S (summary expert) and RO-LMM-P (plan expert). 

Specifically, we train a summary expert using collected raw clinical report and summary notes. During inference, the summary expert receives raw clinical reports, just like in the training scenario. However, for the plan expert, there is a discrepancy between the training and inference scenarios. In other words, we use the training set made up of collected summary notes instead of generated notes from the summary expert, mainly due to cost concerns and the inherent nature of our framework, which generates output sequentially as illustrated in Figure~\ref{intro}.
However, at the inference phase, our model takes the generated notes from the trained summary expert. 
To deal with the input domain differences for the training and inference time,  Noisy Embedding Fine-Tuning (NEFTune)~\cite{jain2024neftune} which inject uniform noise into embedding could be an effective naive solution to handle noisy inputs in this task. 
However, a crucial consideration arises from the nature of the generated notes, since some of them may lie closer to clean inputs (collected notes) and the others deviate towards noisy inputs. To address this, it is essential to train the model to handle both clean and noisy inputs. To preserve the robustness facilitated by NEFTune while enforcing consistency between the prediction given clean and noisy inputs, we introduce Consistency Embedding Fine-Tuning (CEFTune), resulting in RO-LMM-P++\footnote{`+' , `++' denotes the adoption of NEFTune, and CEFTune, respectively}. More details are as follows.
\subsection{Consistency Embedding Fine-Tuning.} \label{sec:CEFT}
To enhance the performance of instruction fine-tuning, NEFTune  ~\cite{jain2024neftune} 
involves injecting a random noise vector into embeddings during the training process as follows:
\begin{align}
\mathcal{L}_{\text{NEFTune}}(\theta)  = \mathbb{E}_{{(\mbx,\mby) \sim D}}\mathcal{L}_{\text{ce}}(f_{\theta}(\mbtx),\mby), \nonumber \\ 
\mbtx = \mbx + (\alpha / \sqrt{LC}) \boldsymbol{\epsilon,\quad \epsilon} \sim \mathcal{U}(-1,1)
\end{align}
 where $\mbx \in \mathbb{R}^{B \times L \times C}$ is the embedding of data, $B$ denotes batch size, $L$ is 
token length, $C$ is embedding dimension, $\alpha$ is a tunable parameter. \add{The perturbed embedding is denoted as $\tilde{\mathbf{x}}$, and $\epsilon$ represents a random noise vector, drawn from a uniform distribution $\mathcal{U}(-1, 1)$. Additionally, $\mathbb{E}$ represents the expectation.} $f_{\theta}(\cdot)$ represents the model parameterized by $\theta$, $\mby$ is the label of text sample. They demonstrated that the effectiveness of incorporating a random noise vector adjusted based on token length, which yields robust results in most fine-tuning scenarios. 

To bolster the model's capability in handling both noisy and clean inputs {for \add{radiotherapy strategy} suggestion task}, we incorporate a regularization loss to encourage consistency as follows:
\begin{eqnarray}
\mathcal{L}_{\text{CEFTune}}(\theta) & = &  \mathcal{L}_{\text{NEFTune}}(\theta) + \lambda \mathcal{R}(\theta, \theta^{-}), \\ 
\text{where} \, \mathcal{R}(\theta, \theta^{-}) & = & d(\mathcal{T}(f_{\theta}(\mbtx)), \mathcal{T}(f_{\theta^-}(\mbx)))  \nonumber
\label{eq:ceft}
\end{eqnarray}
where $\theta^-$ represents the model with stopped gradient, and $d(\cdot,\cdot)$ is used to quantify the discrepancy between $f_{\theta}(\mbtx)$ and $f_{\theta^-}(\mbx)$.  The metric function $d$ can be any capable of measuring distance in a specific space. The function $\mathcal{T}$ performs detokenization to generate sentences.  By introducing the consistency in
the detokenized domain through $\mathcal{T}$,
our objective is to preserve the robustness property of NEFTune while introducing semantic similarity through the integration of consistency between the output of noisy inputs and clean inputs. Since solely minimizing distance in the same embedding space may weaken the robustness and pose limitations on enforcing semantic similarity,
to capture textual similarity between sentences, another contribution of our work is to calculate the distance in the feature space of SentenceBERT~\cite{reimers2019sentence}. \textit{i.e.,} $d(\mathcal{T}(x),\mathcal{T}(y)) = 1 - s(\mathcal{T}(x))^{\top}s(\mathcal{T}(y))/||s(\mathcal{T}(x))||_{1} ||s(\mathcal{T}(y))||_{1}$, where $s(\cdot)$ is the projection by using a pretrained model such as SentenceBERT~\cite{reimers2019sentence}. 
\begin{figure*}[!ht]
\centering
\includegraphics[width=0.8\linewidth]{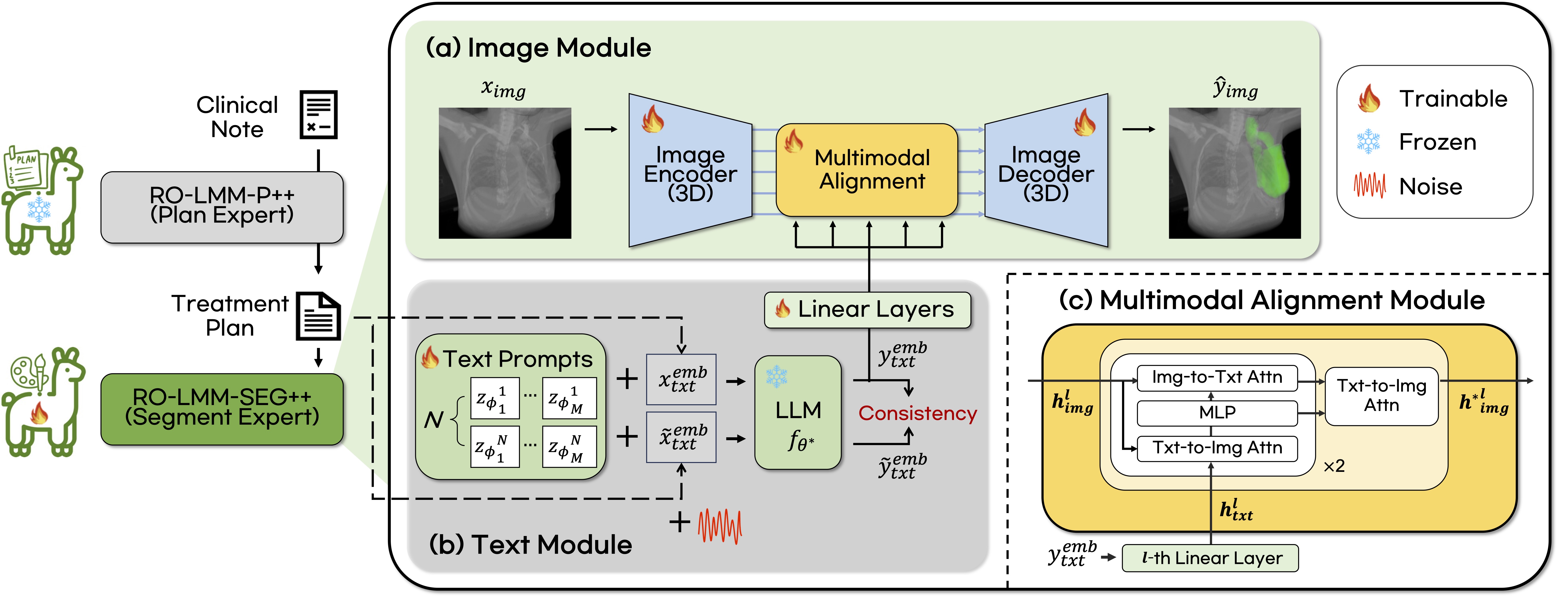}
\caption{Schematics of RO-LMM-SEG++ for plan-guided 3D target volume segmentation task, which composed of (a) image module and (b) text module. These module outputs are aligned through (c) multimodal alignment module.
}
\vspace{-1em}
\label{fig_seg_schematic}
\end{figure*}

\subsection{LMM-assisted 3D Target Volume Segmentation} \label{sec:NESEG}
For incorporating the textual information into target volume segmentation framework, we expand the concept of prompt tuning of LLM for context-aware 3D segmentation introduced in LLMSeg\cite{oh2023llm}. 
{LLMSeg demonstrated significant performance improvement in multimodal radiotherapy target segmentation compared to unimodal models. Additionally, it showed the ability to align target contours with provided clinical text information.}
{However, in the extended clinical workflow proposed here, the LMM-assisted segmentation model receives noisy text descriptions of the generated \add{radiotherapy strategy} from the previous summarization step.
To address this noise, we introduce two novel modules: Noise Embedding Segmentation (NESEG) and Regularized Consistency Embedding Segmentation (CESEG). These modules enhance the target volume segmentation network by handling noise and promoting consistency.} The details are as follows.
\subsection{Noise Embedding Segmentation (NESEG).} 
Firstly, we introduce the noise augmentation technique, dubbed as NESEG, for the target volume segmentation task.  We refer to the corresponding RO-LMM as RO-LMM-SEG+.
Specifically, by extending aforementioned idea of NEFTune, we inject a random noise into input data embeddings $\mbx$ to improve the segmentation model robustness for the generated noisy plan, leading to the follow loss function: 
\begin{align}
\mathcal{L}_{\text{NESEG}}(\Theta)  &= \mathbb{E}_{(\mbix,\mbtx,\mbiy)\sim D} \mathcal{L}_{\text{ce}}(g_{\psi}(\mbix; \mathbf{\mbty} ),\mbiy) \nonumber \\
&\text{where} \; \mbty(\phi) = f_{\theta^*}(\mbtx;\mathbf{z}_{\phi})\, 
\end{align}
where $\Theta$ = [$\psi, \phi$], $\mbix \in \mathbb{R}^{B \times H W S}$ is the 3D CT scan, $B$ denotes batch size, $H$, $W$, and $S$ correspond to height, width, and slice of the CT scan, $\mbty \in \mathbb{R}^{B \times 1 \times C}$ is the perturbed text embedding, $\mbiy \in \mathbb{R}^{B \times H W S}$ is the 3D ground truth segmentation mask, $g_{\psi}(\cdot)$ represents the multimodal segmentation model parameterized by $\psi$, $f_{\theta^*}$ denotes a frozen LLM, $\mbtx$ is the noise injected plan embeddings, and $\mathbf{z}_{\phi}$ is the learnable text prompt parametreized by $\phi$.
\subsection{Consistency Embedding Segmentation (CESEG).} \label{sec:CESEG}
We further adapt the consistency regularization module, dubbed as CESEG,  for generalizing our segmentation model to both the generated noisy plan and the clean ground truth plan. 
The schematic of our multimodal model utilizing CESEG, denoted as RO-LMM-SEG++, is illustrated in Figure~\ref{fig_seg_schematic}.

Apart from the original concept of CEFTune, CESEG modifies the consistency regularization module for the multimodal segmentation task, by combining CEFTune and the text prompt tuning. Once the text prompt-prepended noisy embeddings of \add{radiotherapy strategy} $\mbtx$ are inputted to the multimodal model, the output embedding is also regularized with respect to the clean embeddings $\mbx$. Specifically, the loss function for CESEG is formulated as:
\begin{eqnarray}\label{eq:ceseg}
\mathcal{L}_{\text{CESEG}}(\Theta)  &=& \mathcal{L}_{\text{NESEG}}(\Theta) + \lambda \mathcal{R}(\phi,\phi^-) \\
\text{where} \,\mathcal{R}(\phi,\phi^-) &=& d(\mbty(\phi), \mbey(\phi^-))  \nonumber
\end{eqnarray}
where, $\phi^-$ represents the learnable text prompts with stopped gradient, $d(\cdot,\cdot)$ denotes the cosine similarity, and $\mbey = f_{\theta^*}(\mbx;\mathbf{z}_{\phi^-})\in \mathbb{R}^{B \times 1 \times C}$ is the text embedding for the clean plan $\mbx$.

 We employ the 3D Residual U-Net \cite{cciccek20163d} as for the backbone structure within the image module, as shown in (a). In the text module, as shown in (b), we employ pre-trained RO-LMM-P++ as for the LLM module $f_{\theta^*}$ for producing text embedding $\mbey$ given the \add{radiotherapy strategy} embedding $\mbx$ as input. In the multimodal alignment module, as illustrated as (c), we adopt both self-attention and cross-attention mechanisms in an interactive manner (image-to-text and text-to-image) for each $l$-th skip-connection layer, by following the concept of promptable segmentation borrowed from Segment Anything Model (SAM) \cite{kirillov2023segment}). For realizing multimodal alignment within the 3D segmentation framework, we adapt the idea of light-weight text prompt tuning to transfer the linguistic capability of LLM, by following the concepts introduced in~\cite{oh2023llm, kim2024otseg, zhou2022conditional}. Therefore, entire parameters of the LLM module $f_{\theta^*}$ are frozen, while the learnable text prompts $\mathbf{z}_{\phi}$ and $l$-th linear layers are exclusively optimized within the text module.

\begin{table}[!t]
  \centering
  \caption{Training data details.  CRS: Clinical Report Summarization. \add{RSS:} \add{Radiotherapy Strategy} Suggestion. PTS: Plan-guided Target Segmentation.  US: Ultrasound. Path: Pathology. }
  \resizebox{1\linewidth}{!}{
    \begin{tabular}{lllccc}
    \toprule
    \multirow{2}{*}{\shortstack[c]{\bf{Task}}} & \multirow{2}{*}{\shortstack[c]{\bf{Input}}} & \multirow{2}{*}{\shortstack[c]{\bf{Output}}} &\multicolumn{2}{c}{\textbf{Internal}} & \multicolumn{1}{c}{\textbf{External}}\\
     \cmidrule(lr){4-5}  \cmidrule(lr){6-6} 
     
     & & & \textbf{Train} & \textbf{Validation} & \textbf{Validation} \\
    \midrule
    \bf{CRS   } &  {MRI $\&$ US $\&$ Path Reports           } &  Clinical Note &\multirow{2}{*}{5,674}  & \multirow{2}{*}{\add{121}} & \multirow{2}{*}{81} \\
    \bf{\add{RSS}} & Clinical Note & {\add{Radiotherapy Strategy}        } &    &   \\
    \cmidrule(l){1-1} \cmidrule(l){2-2} \cmidrule(l){3-3} \cmidrule(lr){4-5}  \cmidrule(l){6-6} 
    \bf{PTS} & {\add{Radiotherapy Strategy} $\&$ CT Scan        } & Segmentation & \add{593} & 79  & 81 \\
    
    \bottomrule
    
    \end{tabular}%
    }
  \label{tab_dataset}%

\end{table}%

\section{Experimental settings}
\subsection{Dataset}

To train our model, we collected an internal dataset consisting of 5,674 breast cancer patients treated at the Department of Radiation Oncology at Yonsei Cancer Center. For training the plan-guided 3D target volume segmentation network, we utilized multimodal data from 593 patients, including their 3D CT scans and corresponding ground truth masks.
\add{For the text-related dataset, we established a validation set with 121 patients. Due to the limited availability of image data compared to text data, the validation size for the segmentation task was set to 79 patients.}

To evaluate the model performance on cross-centre datasets, we further acquired external data cohort composed of 81 patients treated at the Department of Radiation Oncology at Yongin Severance Hospital. 
The detail of dataset is described in Table~\ref{tab_dataset} and \ref{sec:appn_dataset}.
We included patients who received their initial diagnosis of breast cancer and subsequently underwent radiation therapy following curative surgery, while excluding individuals with recurrent or metastatic breast cancer in both hospitals. 
This study was approved by the Institutional Review Board (IRB) of each participating hospitals.

\begin{table*}[!t]
    \caption{Quantitative comparison for clinical note summarization. Vanilla:  the instruction fine tuning. CI: confidence interval.} 
    \centering
    \resizebox{.95\linewidth}{!}{
    \begin{tabular}{lccccccc}
    \toprule
      \multirow{2}{*}{\shortstack[c]{\bf{Model}}}   & \multirow{2}{*}{\shortstack[c]{\bf{Method}}}&\multicolumn{3}{c}{\bf{Internal Validation (\add{N=121})}}  & \multicolumn{3}{c}{\bf{External Validation (N=81)}}  \\ 
       \cmidrule(l){3-5} \cmidrule(l){6-8}
       & & Rouge 1 $\uparrow$ (CI) & Rouge 2 $\uparrow$ (CI)& Rouge L $\uparrow$ (CI)& Rouge 1 $\uparrow$ (CI)& Rouge 2 $\uparrow$(CI) & Rouge L $\uparrow$  (CI)\\
    \midrule
    \multicolumn{3}{l}{\bf{Clinical Report Summarization}}\\
         \cmidrule(l){1-1} \cmidrule(l){2-2} \cmidrule(l){3-5} \cmidrule(l){6-8}
         LLaMA2~\cite{touvron2023llama2}  &-& \add{0.621 (0.601-0.642)} & \add{0.434 (0.419-0.449)} & \add{0.621 (0.601-0.642)} & 0.667 (0.649-0.684) & 0.468 (0.456-0.479) & 0.667 (0.649-0.684) \\
         MedAlpaca~\cite{han2023medalpaca}  &-& \add{0.514 (0.460-0.549)} & \add{0.351 (0.318-0.392)} & \add{0.514 (0.460-0.549)} & 0.414 (0.362-0.462) & 0.329 (0.283-0.373) & 0.414 (0.362-0.462)   \\
         ChatDoctor~\cite{li2023chatdoctor}  &-& \add{0.601 (0.544-0.641)} & \add{0.400 (0.370-0.424)} & \add{0.601 (0.544-0.641)} & 0.549 (0.508-0.588) & 0.425 (0.392-0.459) & 0.549 (0.508-0.588) \\ 
         Asclepius~\cite{kweon2023publicly}  &-& \add{0.652 (0.624-0.674)} & \add{0.442 (0.429-0.452)} & \add{0.652 (0.624-0.674)} & 0.590 (0.571-0.609) & 0.412 (0.401-0.423) & 0.590 (0.571-0.609) \\
         PMC-LLaMA~\cite{wu2023pmc}  &-& \add{0.390 (0.334-0.436)} & \add{0.233 (0.194-0.270)} & \add{0.390 (0.334-0.436)} & 0.382 (0.327-0.434) & 0.235 (0.197-0.275) & 0.382 (0.327-0.434) \\
         ChatGPT  &-& \add{0.657 (0.614-0.685)} & \add{0.456 (0.434-0.473)} & \add{0.657 (0.614-0.685)} & 0.651 (0.615-0.683) & 0.456 (0.434-0.476) & 0.651 (0.615-0.683) \\
         \cmidrule(l){1-1} \cmidrule(l){2-2} \cmidrule(l){3-5} \cmidrule(l){6-8}
         RO-LMM-S  & Default & \add{\bf{0.783 (0.758-0.806)}} & \add{\bf{0.605 (0.582-0.625)}} & \add{\bf{0.783 (0.758-0.806)}} & \bf{0.788 (0.769-0.806)} & \bf{0.607 (0.586-0.628)} & \bf{0.788 (0.769-0.806)} \\

    \midrule
    \multicolumn{3}{l}{\bf{\add{Radiotherapy Strategy Suggestion}}}\\
    \cmidrule(l){1-1} \cmidrule(l){2-2} \cmidrule(l){3-5} \cmidrule(l){6-8}
         LLaMA2~\cite{touvron2023llama2}  &-& \add{0.075 (0.068-0.082)} & \add{0.036 (0.032-0.040)} & \add{0.075 (0.068-0.082)} & 0.069 (0.065-0.073) & 0.029 (0.027-0.031) & 0.069 (0.065-0.073) \\
         MedAlpaca~\cite{han2023medalpaca}  &-& \add{0.133 (0.120-0.145)} & \add{0.051 (0.047-0.055)} & \add{0.133 (0.120-0.145)} & 0.137 (0.123-0.152) & 0.048 (0.044-0.053) & 0.137 (0.123-0.152)  \\
         ChatDoctor~\cite{li2023chatdoctor}  &-& \add{0.205 (0.184-0.225)} & \add{0.079 (0.069-0.087)} & \add{0.205 (0.184-0.225)} & 0.191 (0.178-0.205) & 0.065 (0.060-0.071) & 0.191 (0.178-0.205) \\
         Asclepius~\cite{kweon2023publicly}  &-& \add{0.169 (0.155-0.181)} & \add{0.072 (0.064-0.079)} & \add{0.169 (0.155-0.181)} & 0.142 (0.134-0.149) & 0.055 (0.051-0.059) & 0.142 (0.134-0.149) \\
         PMC-LLaMA~\cite{wu2023pmc}  &-& \add{0.113 (0.096-0.131)} & \add{0.033 (0.028-0.038)} & \add{0.113 (0.096-0.131)} & 0.165 (0.136-0.195) & 0.050 (0.040-0.060) & 0.165 (0.136-0.195) \\
         GPT4.0 &-& \add{0.356 (0.326-0.375)} & \add{0.222 (0.197-0.240)} & \add{0.356 (0.326-0.375)} & 0.316 (0.294-0.339) & 0.249 (0.226-0.272) & 0.316 (0.294-0.339) \\         
        \cmidrule(l){1-1} \cmidrule(l){2-2} \cmidrule(l){3-8}
      RO-LMM-P & Default & \add{0.634 (0.590-0.670)} & \add{0.471 (0.418-0.521)} & \add{0.634 (0.590-0.670)} & 0.591 (0.574-0.609) & 0.426 (0.403-0.452) & 0.591 (0.574-0.609) \\
     RO-LMM-P+ & NEFT & \add{0.623 (0.576-0.660)} & \add{0.453 (0.399-0.510)} & \add{0.623 (0.576-0.660)} & 0.588 (0.573-0.606) & 0.419 (0.399-0.441) & 0.588 (0.573-0.606) \\
     RO-LMM-P++&  CEFT & \add{\bf{0.655 (0.616-0.693)}} & \add{\bf{0.500 (0.449-0.554)}} & \add{\bf{0.655 (0.616-0.693)}} & \bf{0.615 (0.592-0.638)} & \bf{0.459 (0.429-0.489)} & \bf{0.615 (0.592-0.638)} \\

    \bottomrule
    \end{tabular}
    }
    \label{tab_summary}

\end{table*}

\begin{table}[!hbt]
    \caption{Clinical expert analysis for report summarization. R\#: each rubric,  C\#:each clinical expert.} 
    \centering
    \resizebox{1\linewidth}{!}{
    \begin{tabular}{ccccc}
          \toprule
         & \multicolumn{3}{c}{\shortstack[c]{\bf{Internal Validation (N=10) }}} \\
         \cmidrule(lr){2-5} 
            & \bf{ChatGPT} & \bf{LLaMa2-7B}  & \bf{RO-LMM-S}  \\
        \cmidrule(lr){2-2}  \cmidrule(lr){3-3}  \cmidrule(lr){4-4}  \cmidrule(lr){5-5}      
         \bf{Experts} & R1 / R2 / R3 / R4 / Total & R1 / R2 / R3 / R4 / Total & R1 / R2 / R3 / R4 / Total  & $r$\\          
        \cmidrule(l){1-1} \cmidrule(l){2-2}  \cmidrule(lr){3-3}  \cmidrule(lr){4-4} \cmidrule(lr){5-5} 
         C1 &  3.0 /	 0.0 /	4.0 / 7.0 / 13.3	&	2.0 / 0.0 /	0.0 /	7.0	/ 9.0  &	15.0 /	10.0	/	10.0	/	10.0	/	\bf{45.0} & \multirow{2}{*}{0.884} \\
         C2 & 3.0 / 0.0 / 5.0 / 7.7 / \add{15.4} & 1.0 / 0.0 / 0.0 / 9.0 / 10.0 & 15.0 / 10.0 / 9.0 / 10.0 / \bf{44.0} \\
                \cmidrule(l){1-1} \cmidrule(l){2-2}  \cmidrule(lr){3-3}  \cmidrule(lr){4-4} \cmidrule(lr){5-5} 
                
         & \multicolumn{3}{c}{\shortstack[c]{\bf{External Validation (N=10) }}} & $r$  \\
        \cmidrule(l){1-1} \cmidrule(l){2-2}  \cmidrule(lr){3-3}  \cmidrule(lr){4-4} \cmidrule(lr){5-5} 
         C1 &  5.0 /	 1.0 /	9.0 /	7.0	/ 22.0	&	7.0	/	0.0	/ 0.0 /	7.0	/ 14.0 &	10.0 /	9.0	/	8.0	/	8.0	/	\bf{35.0}	&	\multirow{2}{*}{0.977}\\
         C2&  6.0	/	3.0	/	9.0 /	\add{7.0}	/	25.0	&	8.0	/	0.0	/	0.0	/	9.0	/	17.0	&	10.0	/	9.0	/	8.0	/	9.0	/	\bf{36.0} \\ 
    \bottomrule 

    \end{tabular}
    }
    \label{tab1compare_summary}

\end{table}

\subsection{Implementation details}
For instruction fine-tuning for RO-LMM-S and RO-LMM-P, we use the LLaMA-2-7B-Chat~\cite{touvron2023llama2} model as a Default. 
\add{We adopt full-finetuning approach for all of our models.} The maximum context length is set at 4096, and the batch size is set to 2 for each GPU, leveraging 4 NVIDIA A6000 GPUs. We use the AdamW~\cite{loshchilov2017decoupled} optimizer, with a learning rate of 5e-5 until reaching 3 epochs for both RO-LMM-S and RO-LMM-P. The hyper-parameter $\lambda$ for CEFTune is set to 1 in all of tasks, and for consistency regularization we adopt the variants of SBERT~\cite{reimers2019sentence} which is trained on PubMed\footnote{\url{https://www.ncbi.nlm.nih.gov/pubmed/}} dataset, called PubMedBERT.\footnote{\url{www.huggingface.co/NeuML/pubmedbert-base-embeddings}.} \add{To ensure fair comparisons with other approaches, we utilize the Fully Sharded Data Parallel (FSDP) acceleration method, the bf16 (Brain Floating Point) data format, and gradient checkpointing~\cite{chen2016training} in all experiments.} 

\add{For training the segmentation network, we adopt the 3D Residual U-Net \cite{cciccek20163d} as the backbone architecture, implemented using the MONAI open-source library\footnote{\url{https://monai.io/}}. During data preprocessing, all chest CT scans and CTV annotations are re-sampled to a uniform voxel spacing of 1.0 $\times$ 1.0 $\times$ 3.0 mm$^3$. Image intensities are truncated to a Hounsfield Unit (HU) range of -1,000 to 1,000 and normalized linearly to fall within a scale of 0 to 1.0. For network training, 3D patches sized 384 $\times$ 384 $\times$ 128 pixels are randomly cropped to encompass the entire breast, using a batch size of 1 per GPU. Training utilizes 4 NVIDIA A6000 GPUs. Performance of the backbone is benchmarked against state-of-the-art models such as 3D SegMsamba \cite{xing2024segmamba} and 3D UNETR \cite{hatamizadeh2022unetr}. Due to memory constraints, input sizes of 384 $\times$ 384 $\times$ 128 pixels and 320 $\times$ 320 $\times$ 64 pixels are used for SegMamba and UNETR, respectively. For evaluation, the complete 3D CT volumes are processed using a sliding window approach across all models. To incorporate textual information into the RO-LMM-SEG/SEG+/SEG++ frameworks, we initialize the LLaMA-2-7B-Chat model with the pre-trained RO-LMM-P++ checkpoint. For training all the segmentation models, optimization is performed with the AdamW optimizer \cite{loshchilov2017decoupled}, using a learning rate of 1e-4 over 100 epochs. }

\begin{table*}[!t]
    \caption{Clinical expert analysis for \add{radiotherapy strategy} suggestion. R\#: each rubric, C\#: each clinical expert.} 
    \centering
    \resizebox{0.9\linewidth}{!}{
    \begin{tabular}{cccccc}
          \toprule
         & \multicolumn{4}{c}{\shortstack[c]{\bf{Internal Validation (N=25) }}} \\
         \cmidrule(lr){2-6} 
            & \bf{GPT-4.0} & \bf{RO-LMM-P}  & \bf{RO-LMM-P+ (NEFT)}  & \bf{RO-LMM-P++ (CEFT)}  \\
        \cmidrule(lr){2-2}  \cmidrule(lr){3-3}  \cmidrule(lr){4-4}  \cmidrule(lr){5-5} \cmidrule(lr){6-6}     
         \bf{Experts} & R1 / R2 / R3 / R4 / R5  / Total & R1 / R2 / R3 / R4 / R5 / Total & R1 / R2 / R3 / R4 / R5 / Total & R1 / R2 / R3 / R4 / R5 / Total & $r$\\ 
         
        \cmidrule(l){1-1} \cmidrule(l){2-2}  \cmidrule(lr){3-3}  \cmidrule(lr){4-4} \cmidrule(lr){5-5} \cmidrule(lr){6-6}
         C1 & 8.8	/	8.4	/	4.4	/	7.2	/	2.0	/	30.8	&	5.6	/	10.0	/	8.8	/	9.2	/	9.2	/	\bf{42.8}	&	5.6	/	9.6	/	8.8	/	9.2	/	9.2	/	42.4	&	4.8	/	10.0	/	8.8	/	9.2	/	10.0	/	\bf{42.8} & \multirow{2}{*}{0.7} \\
         C2 & 9.6 / 9.6 / 5.6 / 4.4 / 3.8 / 32.7 & 5.6 / 10.0 / 8.4 / 8.8 / 9.2/ \bf{42.0} & 5.6 / 9.6 / 8.0 / 8.4/ 10.0 / 41.6 & 4.4 / 10.0 / 8.8 / 9.2 / 9.6 / \bf{42.0} & \multirow{2}{*}{0.4}\\
         GPT-4.0 & 10.0 / 7.6 / 4.8 / 5.6 / 6.8 / 29.2 & 8.2 / 8.0 / 7.2 / 8.8 / 9.6 / 37.2 & 9.1 / 7.6 / 6.4 / 7.6 / 9.2 / 34.8 &  9.1 / 8.4 / 8.0 / 8.8 / 10.0 / \bf{39.2} \\
                \cmidrule(l){1-1} \cmidrule(l){2-2}  \cmidrule(lr){3-3}  \cmidrule(lr){4-4} \cmidrule(lr){5-5} \cmidrule(lr){6-6} 
                
         & \multicolumn{4}{c}{\shortstack[c]{\bf{External Validation (N=25) }}} & $r$  \\
        \cmidrule(l){1-1} \cmidrule(l){2-2}  \cmidrule(lr){3-3}  \cmidrule(lr){4-4} \cmidrule(lr){5-5} \cmidrule(lr){6-6}
         C1 &  8.5	/	8.8	/	7.3	/	6.5	/	1.2	/	32.3	&	8.1	/	8.8	/	6.9	/	6.9	/	9.2	/	40.0	&	7.7	/	9.6	/	7.7	/	7.3	/	9.6	/	41.9	&	8.5	/	9.6	/	8.8	/	9.6	/	9.6	/	\bf{46.2} & \multirow{2}{*}{0.9}\\
         C2&  8.8	/	8.5	/	6.2	/	3.4	/	7.3	/	34.2	&	8.5	/	9.6	/	7.7	/	7.3	/	9.2	/	42.3	&	8.1	/	9.6	/	8.1	/	8.8	/	9.2	/	43.8	&	8.5	/	9.6	/	8.8	/	8.8	/	9.6	/	\bf{45.8}  & \multirow{2}{*}{0.4}\\ 
         GPT-4.0 & 	8.8 / 6.9 / 6.2 / 5.8 / 4.2 / 31.9	&8.1 / 7.3 / 6.2 / 1.5 / 8.8 / 31.9	&8.5 / 8.1 / 5.0 / 1.2 / 9.2 / 31.9	&	8.5 / 8.1 / 6.5 / 3.8 / 9.2 / \bf{36.2}\\ 

    \bottomrule 

    \end{tabular}
    }
    \label{tab_compare}
\end{table*}

\subsection{Evaluation Metric and Default Models} \label{eval}
The evaluation of generated clinical report summaries and \add{radiotherapy strategies} utilizes common NLP metrics, including Rouge-1 (R-1), Rouge-2 (R-2), and Rouge-L (R-L)~\cite{lin2004rouge}.

However, acknowledging potential limitations in domain specificity for both tasks, 
{we conduct evaluations using two clinical experts and GPT-4.0. A board-certified expert creates five scoring rubrics for plan suggestions, detailed in Table~\ref{table_prompt}. Using these rubrics, the experts score each aspect of the plans, assigning 0 or 1 for each rubric separately. Similarly, for clinical report summarization, rubrics are created and the same evaluation process is followed.} Inspired by the recent evaluation methodologies such as GPT-Score~\cite{fu2023gptscore} and G-Eval~\cite{liu2023g}, GPT-4.0 also evaluates specific examples of outputs based on these rubrics with in-context learning on the \add{radiotherapy strategy} suggestion task.
To evaluate the 3D target volume segmentation performance using a ground truth segmentation mask, we measure Dice coefficient (Dice), Intersection over Union (IoU), and the 95th percentile of Hausdorff Distance (HD-95) \cite{crum2006generalized} in \add{millimeter (mm)} unit to measure spatial distances between the ground truth masks and the predicted segmentation results. 

To further validate effectiveness of our experiment, we provide the confidence intervals (CI) on each metric by utilizing the non-parametric bootstrap method. We calculate CI by conducting 1,000 times of random sampling procedure with replacement from the experimental results, each time equal to the total size of the sample dataset. Subsequently, we estimate the 95th percentile CI from the relative frequency distribution of each experimental trial.

For text-related tasks, to validate the effectiveness of our model, we compare our model with several clinical LLMs serving as Defaults, specifically MedAlpaca~\cite{han2023medalpaca}, ChatDoctor~\cite{li2023chatdoctor}, Asclepius~\cite{kweon2023publicly}, and PMC-LLAMA~\cite{wu2023pmc}. Additionally, we include a comparison with ChatGPT~\cite{chatgpt} for the report summarization task and GPT-4.0 for the \add{radiotherapy strategy} suggestion task using few-shot in-context learning.

For the segmentation task, we utilize 3D Residual U-Net~\cite{cciccek20163d} and ConTEXTualNET~\cite{huemann2023contextual} as our Default methods. We further utilize an additional Default as a variant of RO-LMM-SEG, namely LLaMA2-SEG, which is initialized with the checkpoint of vanilla LLaMA-2-7B-Chat model.

\section{Results}
\label{sec:result}

\subsection{Clinical Report Summarization.}

{We present the performance of our model on the clinical report summarization task, along with confidence intervals for each method, in Table~\ref{tab_summary}. Our fine-tuned model of RO-LMM-S demonstrate significant improvements over the Defaults, providing consistent margins in all metrics and confidence intervals. Notably, RO-LMM-S outperforms ChatGPT with few-shot in-context learning.}

Moreover, we evaluate the generated summaries using expertise-based rubrics by two clinical experts and compare them to Defaults, including ChatGPT and LLaMa-2. As shown in Table \ref{tab1compare_summary}, our RO-LMM-S model significantly outperforms all Defaults in both internal and external validations, thanks to its domain-specific knowledge. Additionally, Pearson correlation ($r$) analysis reveals strong positive inter-clinician correlations ( $> 0.85$ and  $> 0.95$ for internal and external validation, respectively), confirming the reliability of our rubrics and the clinical relevance of RO-LMM-S.  Therefore, our RO-LMM-S provides practical and meaningful summaries that can assist in the field of radiation oncology.

\subsection{\add{Radiotherapy Strategy} Suggestion.}
As indicated in Table~\ref{tab_summary}, we evaluate the model's performance on \add{radiotherapy strategy} suggestion task based on generated clinical notes. Similar to the report summarization task, our fine-tuned variants, RO-LMM-P, exhibit substantial enhancements in suggestion performance compared to other Defaults. Notably, RO-LMM-P+ shows inferior performance on the external dataset compared to the vanilla approach. However, our proposed RO-LMM-P++ achieves the best performance on both internal and external datasets across all metrics.

{Furthermore, we comprehensively evaluate the generated plans using our expertise-based rubrics and clinical expert analysis, as detailed in Table \ref{tab_compare}. For both internal and external validation, our RO-LMM variants significantly outperform GPT-4.0 with few-shot in-context learning, as assessed by both GPT and clinical experts, due to their domain-specific knowledge.}
{We confirm that our model with CEFT achieved the best performance in terms of clinical evaluation, especially on the external set, demonstrating clinical effectiveness in line with our goals. We further evaluate inter-clinicians correlation and the correlation between clinicians and GPT-4, using Pearson correlation ($r$). The results show strong positive correlations ($>0.7$ and $>0.4$, for each), affirming reliability of our proposed evaluation rubrics.}

\begin{figure*}[!ht]
    \centering
    \begin{minipage}[t]{1\textwidth}
        \centering
        \includegraphics[width=1\linewidth]{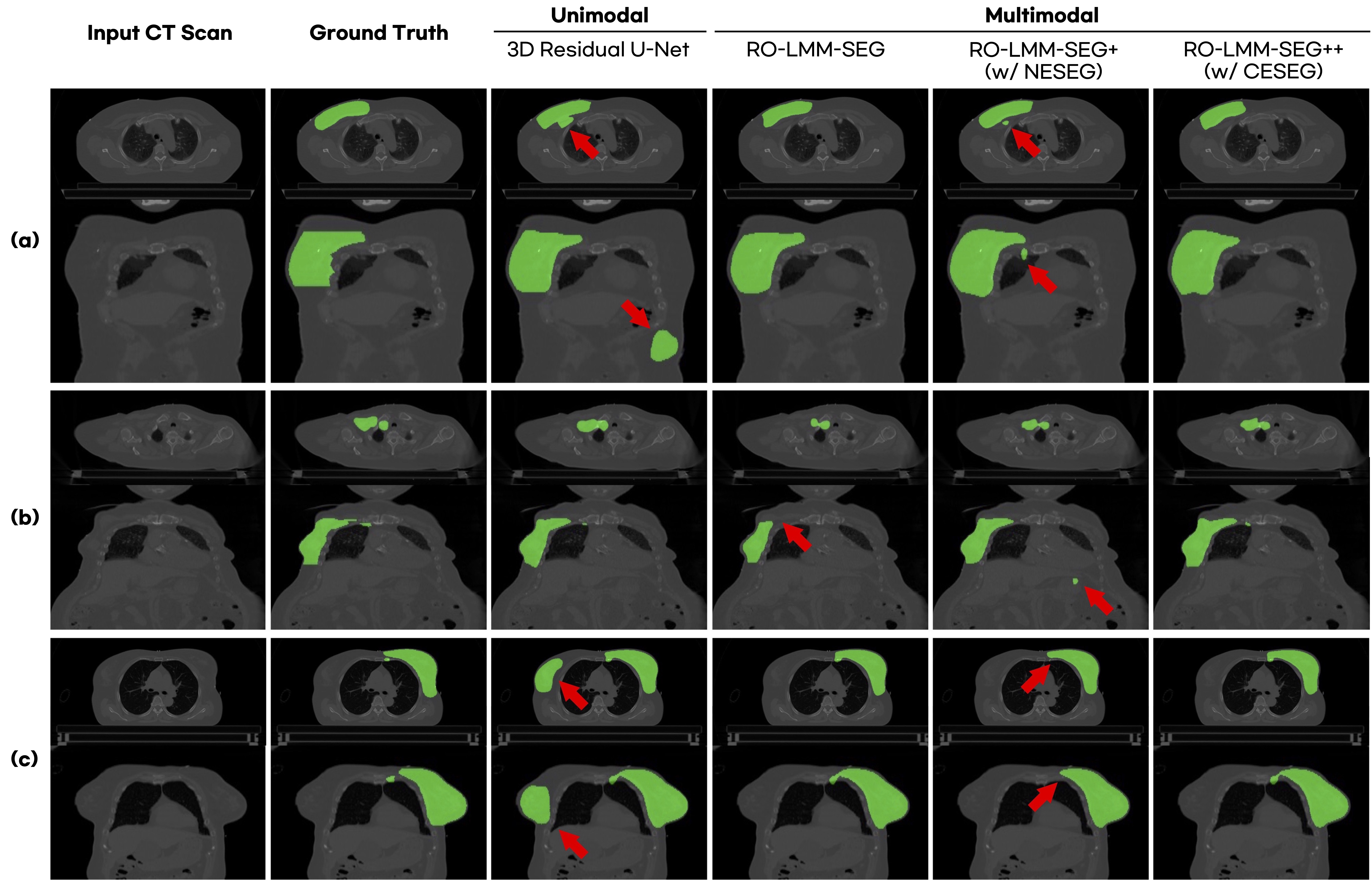}
        \caption{Qualitative comparison on 3D target volume segmentation task. Red arrows indicate errors.}
        \label{fig_seg_main}
    \end{minipage}
\end{figure*}

\begin{table*}[!t]
    \caption{Comparison of 3D target volume segmentation performance.} 
    \centering
    \resizebox{1\linewidth}{!}{
    \begin{tabular}{lcccccccccc}
    
      \toprule
      \multirow{2}{*}{\shortstack[c]{\bf{Model}}} &  \multirow{2}{*}{\shortstack[c]{\\\bf{LMM}\\\bf{Tuning}}} & \multicolumn{4}{c}{\shortstack[c]{\bf{Internal Validation (N=79)}}}  & \multicolumn{4}{c}{\shortstack[c]{\bf{External Validation (N=81)}}}    \\  

        \cmidrule(l){3-6} \cmidrule(l){7-10}
        
        & & Dice $\uparrow$ (CI) & \add{surfaceDice} $\uparrow$ (CI) & IoU $\uparrow$ (CI) &  HD-95 $\downarrow$ (CI)  & Dice $\uparrow$ (CI)  & \add{surfaceDice} $\uparrow$ (CI) & IoU $\uparrow$ (CI) & HD-95 $\downarrow$ (CI)  \\
         \midrule 
         
         {{3D Residual U-Net \cite{cciccek20163d}}} & - & \add{{0.802} {(0.775-0.826)}} & \add{{0.517} {(0.485-0.545)}} & \add{{0.684} {(0.650-0.714)}} & \add{{45.852} {(32.441-61.029)}} & \add{{0.689} {(0.652-0.723)}} & \add{{0.409} {(0.381-0.436)}} & \add{{0.548} {(0.510-0.584)}} & \add{{132.825} {(113.296-151.525)}}   \\
         {\add{3D SegMamba \cite{xing2024segmamba}}} & - &  \add{{0.576} {(0.554-0.598)}} & \add{{0.364} {(0.344-0.386)}} & \add{{0.411} {(0.389-0.434)}} & \add{{146.610} {(140.391-152.103)}} & \add{{0.569} {(0.540-0.595)}} & \add{{0.315} {(0.295-0.334)}} & \add{{0.408} {(0.381-0.434)}} & \add{{156.767} {(152.551-160.849)}} \\
         {\add{3D UNETR \cite{hatamizadeh2022unetr}}} & - & \add{{0.564} {(0.544-0.584)}} & \add{{0.340} {(0.319-0.361)}} & \add{{0.398} {(0.379-0.419)}} & \add{{147.310} {(141.625-152.392)}} & \add{{0.538} {(0.523-0.551)}} & \add{{0.302} {(0.287-0.316)}} & \add{{0.370} {(0.357-0.382)}} & \add{{158.272} {(154.231-162.241)}} \\

        \cmidrule(l){1-1} \cmidrule(l){2-2} \cmidrule(l){3-6} \cmidrule(l){7-10}

        \multirow{1}{*}{{LLMSeg \cite{oh2023llm}}} & - & \add{{0.814} {(0.790-0.836)}} & \add{{0.531} {(0.499-0.561)}} & \add{{0.699} {(0.668-0.727)}} & \add{{42.814} {(29.983-56.439)}} & \add{{0.681} {(0.656-0.700)}} & \add{{0.397} {(0.373-0.420)}} & \add{{0.525} {(0.502-0.544)}} & \add{{160.151} {(150.742-169.029)}}   \\ 
        
        $\text{ConTEXTualNET}^\dag$ \cite{huemann2023contextual} & - & \add{{0.774} {(0.728-0.811)}} & \add{{0.491} {(0.451-0.525)}} & \add{{0.664} {(0.618-0.703)}} & \add{{31.772} {(21.484-44.095)}} & \add{{0.745} {(0.708-0.776)}} & \add{{0.444} {(0.413-0.472)}} & \add{{0.622} {(0.586-0.654)}} & \add{{82.013} {(64.103-100.648)}}  \\

         \cmidrule(l){1-1} \cmidrule(l){2-2} \cmidrule(l){3-6} \cmidrule(l){7-10}

         \multirow{1}{*}{{RO-LMM-SEG}} & \add{Default} & \add{{0.818} {(0.795-0.841)}} & \add{{0.524} {(0.493-0.555)}} & \add{{0.705} {(0.676-0.736)}} & \add{{20.611} {(13.502-29.389)}} & \add{{0.747} {(0.715-0.775)}} & \add{{0.419} {(0.388-0.449)}} & \add{{0.615} {(0.580-0.646)}} & \add{{100.562} {(86.656-115.053)}} 

        \\
         
        \multirow{1}{*}{{RO-LMM-SEG+}} & NESEG  & \add{{0.828} {(0.806-0.850)}} & \add{{0.518} {(0.489-0.547)}} & \add{{0.718} {(0.690-0.745)}} & \add{{17.805} {(10.818-26.427)}} & \add{\bf{0.774} {(0.733-0.811)}} & \add{\bf{0.458} {(0.427-0.486)}} & \add{\bf{0.659} {(0.619-0.695)}} & \add{{34.487} {(23.313-46.962)}}   \\

         \multirow{1}{*}{{RO-LMM-SEG++}} & CESEG & \add{\bf{0.838} {(0.818-0.858)}} & \add{\bf{0.555} {(0.523-0.583)}} & \add{\bf{0.732} {(0.706-0.757)}} & \add{\bf{13.658} {(8.133-20.770)}} & \add{{0.761} {(0.723-0.794)}} & \add{{0.418} {(0.391-0.443)}} & \add{{0.636} {(0.601-0.668)}} & \add{\bf{29.459} {(21.433-38.557)}}  \\

    \bottomrule 

    \multicolumn{8}{l}{$^\dag$modified for 3D target volume segmentation}\\

    \end{tabular}
    }
    \label{tab_seg}
\end{table*}

\begin{table*}[!t]
    \caption{Comparison of 3D target segmentation performance for overall and specific patient types.} 
    \centering
    \resizebox{1\linewidth}{!}{
    \begin{tabular}{llcccccccccc}
    
    \toprule
       \multirow{2}{*}{\bf{Types}} & \multirow{2}{*}{\bf{Method}} & \multicolumn{5}{c}{\shortstack[c]{\bf{Internal Validation}}}  & \multicolumn{5}{c}{\shortstack[c]{\bf{External Validation}}}    \\ 
        \cmidrule(l){3-7} \cmidrule(l){8-12}
        & & \#Patients & Dice $\uparrow$ (CI) & \add{surfaceDice} $\uparrow$ (CI) & IoU $\uparrow$ (CI) & HD-95 $\downarrow$ (CI) & \#Patients & Dice $\uparrow$ (CI) & \add{surfaceDice} $\uparrow$ (CI) & IoU $\uparrow$ (CI) & HD-95 $\downarrow$ (CI)  \\
        \midrule 
        
        \multirow{3}{*}{\bf{All}} & Unimodal & \multirow{3}{*}{79} & \add{{0.802} {(0.775-0.826)}} & \add{{0.517} {(0.485-0.545)}} & \add{{0.684} {(0.650-0.714)}} & \add{{45.852} {(32.441-61.029)}}  & \multirow{3}{*}{81} & \add{{0.689} {(0.652-0.723)}} & \add{{0.409} {(0.381-0.436)}} & \add{{0.548} {(0.510-0.584)}} & \add{{132.825} {(113.296-151.525)}} \\

        & Multimodal & & 
        \add{\bf{0.838} {(0.818-0.858)}} & \add{\bf{0.555} {(0.523-0.583)}} & \add{\bf{0.732} {(0.706-0.757)}} & \add{\bf{13.658} {(8.133-20.770)}} & & \add{\bf{0.761} {(0.723-0.794)}} & \add{\bf{0.418} {(0.391-0.443)}} & \add{\bf{0.636} {(0.601-0.668)}} & \add{\bf{29.459} {(21.433-38.557)}}  \\

        \cmidrule(r){2-2} \cmidrule(r){4-7}  \cmidrule(r){9-12}  
        & Gain $\uparrow$  & & \correct{+0.036} &   \correct{+0.038} &  \correct{+0.048} & \correct{-29.194} & & \correct{+0.072} &  \correct{+0.009} & \correct{+0.088} & \correct{-103.366} \\

        \midrule

        \multirow{3}{*}{\bf{Typical}} & Unimodal & \multirow{3}{*}{55}  & \add{{0.837} {(0.812-0.860)}} & \add{{0.553} {(0.521-0.583)}} & \add{{0.728} {(0.694-0.760)}} & \add{{37.449} {(21.053-55.604)}}   & \multirow{3}{*}{73}   & \add{{0.720} {(0.691-0.745)}} & \add{{0.429} {(0.403-0.454)}} & \add{{0.576} {(0.542-0.606)}} & \add{{133.108} {(113.319-155.072)}}  \\

        & Multimodal &  
         & \add{\bf{0.876} {(0.865-0.888)}} & \add{\bf{0.603} {(0.576-0.630)}} & \add{\bf{0.783} {(0.765-0.799)}} & \add{\bf{7.385} {(6.095-8.708)}} &  & \add{\bf{0.776} {(0.737-0.808)}} & \add{\bf{0.427} {(0.398-0.456)}} & \add{\bf{0.652} {(0.615-0.684)}} & \add{\bf{27.519} {(19.673-36.031)}}  \\

        \cmidrule(r){2-2} \cmidrule(r){4-7}  \cmidrule(r){9-12} 
        & Gain $\uparrow$ & & \correct{+0.039} & \correct{+0.050} &  \correct{+0.055} & \correct{-30.064} & & \correct{+0.056} & \correct{-0.002} & \correct{+0.076} & \correct{-105.589} \\

        \midrule

        \multirow{3}{*}{\bf{Atypical}} & Unimodal & \multirow{3}{*}{24} &  \add{{0.722} {(0.658-0.781)}} & \add{{0.435} {(0.374-0.494)}} & \add{{0.585} {(0.516-0.652)}} & \add{{67.873} {(36.956-98.711)}} & \multirow{3}{*}{8}  & \add{{0.403} {(0.218-0.593)}} & \add{{0.225} {(0.108-0.349)}} & \add{{0.292} {(0.132-0.460)}} & \add{{128.913} {(91.743-157.660)}}  \\

        & Multimodal & &
        \add{\bf{0.758} {(0.702-0.801)}} & \add{\bf{0.447} {(0.394-0.500)}} & \add{\bf{0.624} {(0.562-0.674)}} & \add{\bf{28.125} {(9.343-50.876)}} & & \add{\bf{0.624} {(0.441-0.751)}} & \add{\bf{0.328} {(0.217-0.425)}} & \add{\bf{0.490} {(0.340-0.603)}} & \add{\bf{48.465} {(12.673-92.137)}} \\

        \cmidrule(r){2-2} \cmidrule(r){4-7}  \cmidrule(r){9-12}
        & Gain $\uparrow$ & & \correct{+0.036} & \correct{+0.012} & \correct{+0.039} & \correct{-39.748} & & \correct{+0.221} & \correct{+0.103}& \correct{+0.198} & \correct{-80.448} \\

        \midrule

        \multirow{3}{*}{\bf{Breast-only}} & Unimodal & \multirow{3}{*}{24} &   \add{{0.837} {(0.787-0.878)}} & \add{{0.573} {(0.516-0.627)}} & \add{{0.733} {(0.668-0.790)}} & \add{{8.040} {(4.885-11.685)}}   & \multirow{3}{*}{51}  & \add{{0.739} {(0.705-0.771)}} & \add{\bf{0.452} {(0.423-0.481)}} & \add{{0.599} {(0.559-0.639)}} & \add{{139.910} {(114.241-167.642)}}   \\

        & Multimodal & &
         \add{\bf{0.891} {(0.870-0.908)}} & \add{\bf{0.625} {(0.580-0.669)}} & \add{\bf{0.806} {(0.775-0.833)}} & \add{\bf{2.985} {(2.365-3.616)}}  &  & \add{\bf{0.786} {(0.749-0.820)}} & \add{{0.441} {(0.411-0.473)}} & \add{\bf{0.664} {(0.625-0.700)}} & \add{\bf{29.320} {(20.006-40.850)}} \\

        \cmidrule(r){2-2} \cmidrule(r){4-7}  \cmidrule(r){9-12}
        & Gain $\uparrow$ & & \correct{+0.054} & \correct{+0.052} & \correct{+0.073} & \correct{-5.055} & & \correct{+0.047} & \correct{-0.011} & \correct{+0.065} & \correct{-110.59} \\ 

        \bottomrule 
        
    \end{tabular}
    }\label{tab_seg_atypical}

\vspace{-1em}
\end{table*}

\begin{table}[!h]
    \caption{{Quantitative comparison results for our RO-LMM's clinical report summarization and \add{radiotherapy strategy} suggestion performance on the publicly available dataset.}} 
    \centering
    \resizebox{1\linewidth}{!}{
    \begin{tabular}{lccc}
    \toprule
      \multirow{2}{*}{\shortstack[c]{\bf{Model}}}  &\multicolumn{3}{c}{\bf{Pubilc available dataset (N=15)}}  \\ 
       \cmidrule(l){2-4} 
       & R-1 $\uparrow$ (CI) & R-2 $\uparrow$ (CI)& R-L $\uparrow$ (CI)\\
    \midrule
    \multicolumn{3}{l}{\bf{Clinical Report Summarization}}\\
        \cmidrule(l){1-1} \cmidrule(l){2-4}
         MedAlpaca~\cite{han2023medalpaca}  & 0.597 (0.502-0.683) & 0.431 (0.343-0.516) & 0.597 (0.502-0.683)\\
         ChatDoctor~\cite{li2023chatdoctor}  & 0.644 (0.511-0.757) & 0.476 (0.370-0.573) & 0.644 (0.511-0.757) \\ 
         Asclepius~\cite{kweon2023publicly}  & 0.696 (0.658-0.733) & 0.469 (0.444-0.493) & 0.696 (0.658-0.733) \\
         ChatGPT  & 0.765 (0.733-0.795) & 0.526 (0.498-0.554) & 0.765 (0.733-0.795)  \\
         \cmidrule(l){1-1} \cmidrule(l){2-4} 
         Ours (RO-LMM-S)  & \bf{0.795 (0.746-0.833)} & \bf{0.569 (0.530-0.612)} & \bf{0.795 (0.746-0.833)} \\
    \midrule
    \multicolumn{3}{l}{\bf{\add{Radiotherapy Strategy} Suggestion}}\\
     \cmidrule(l){1-1} \cmidrule(l){2-4} 
         MedAlpaca~\cite{han2023medalpaca}  & 0.165 (0.111-0.233) & 0.085 (0.064-0.109) & 0.165 (0.111-0.233) \\
         ChatDoctor~\cite{li2023chatdoctor}  & 0.189 (0.147-0.233) & 0.104 (0.085-0.122) & 0.189 (0.147-0.233)  \\ 
         Asclepius~\cite{kweon2023publicly}  & 0.125 (0.102-0.151) & 0.074 (0.061-0.089) & 0.125 (0.102-0.151) \\
         GPT-4.0  & 0.390 (0.302-0.485) & 0.231 (0.179-0.292) & 0.390 (0.302-0.485)\\
         \cmidrule(l){1-1} \cmidrule(l){2-4} 
         Ours (RO-LMM-P++) & \bf0.669 (0.588-0.740) &\bf 0.450 (0.374-0.525) & \bf0.669 (0.588-0.740) \\
    \bottomrule
    
    \end{tabular}
    }
    \label{tab_public}
    \vspace{-1em}
\end{table}

\subsection{3D Target Volume Segmentation.}
{As reported in Table~\ref{tab_seg}, \add{we first compare the performance of various 3D segmentation models for target volume segmentation. Notably, both 3D SegMamba \cite{xing2024segmamba} and 3D UNETR \cite{hatamizadeh2022unetr} demonstrate inferior performance compared to the 3D Residual U-Net. Transformer-based or Mamba-based architectures significantly increase computational memory requirements, which constrains the input image dimensions. In the context of radiotherapy segmentation, where the network must process and align volumetric information in the entire volume-level, such input size limitations can be regarded as a significant limitation, especially under constrained computational resources. Consequently, we selected the more efficient 3D Residual U-Net as the backbone to maximize the 3D input dimensions and ensure optimal performance.} Furthermore, our proposed plan-guided 3D target segmentation frameworks (RO-LMM-SEG/SEG+/SEG++) excel other default methods with large gains, specifically remained stable performance across the external validation setting. This can be also observed in qualitative results. As shown in Figure~\ref{fig_seg_main}(a) and (c), despite the ground truth mask posing target volume on the one side of the breast, the unimodal 3D Residual U-Net incorrectly segments contours not only the target volume but also the outside of the target breast, while all the multimodal RO-LMM variants correctly segment target volume aligning consistently with the ground truth masks. 
\add{Moreover, as shown in Figure~\ref{fig_seg_main}(a) and (b), RO-LMM-SEG+ with NESEG module yields noisy segmentation output as indicated as red arrows, and RO-LMM-SEG sometimes under-segments the target volume. In contrast, our proposed RO-LMM-SEG++ with CESEG module accurately contours the breast and regional lymph nodes that need to be treated. Through experiments on consecutive tasks, including radiotherapy strategy suggestion and 3D target volume segmentation, we found that simple noise injection strategies (NEFT and NESEG) are not consistently effective across all tasks. In contrast, our proposed CEFT and CESEG methods demonstrate robustness and offer a more generalizable solution within the ROB-LMM framework.}

We further categorize all patients into two distinct types: patients those who had breast-conserving surgery (typical type) and patients those who underwent total mastectomy (atypical type). The comparative results are presented in the Table~\ref{tab_seg_atypical} separately, with the unimodal model referred to the 3D Residual U-Net and the multimodal model denoted as our proposed RO-LMM-SEG++ with CESEG. Firstly, our proposed multimodal model shows substantial performance improvements of approximately up to 5\% and 10\% for internal and external validation, respectively. When comparing the typical and atypical types of patients, particularly in the less prevalent atypical type during external validation, we observed notable enhancements with a performance gain of up to 20\%. This substantial generalizbility observed in extremely low-prevalence distributions demonstrates the data efficiency of our multimodal concept, leveraging textual information compared to the traditional unimodal approach.  \add{Additionally, we specifically  categorize cases where regional lymph node treatment is not required and the CTV definition includes only the breast in Table~\ref{tab_seg_atypical}, denoted as breast-only, and compare our results with other studies focused on breast segmentation for radiotherapy, as presented in \cite{smine2024automated}. Our method demonstrates comparable performance, achieving a Dice score of 0.891 and an HD-95 of 2.985 mm for the internal dataset.}
} \\

\subsection{Inference on Public Dataset.} 
To validate the effectiveness of our methods when AI-generated results are reused as input, we evaluated the performance by inputting each model's report summarization results its plan suggestion task. We conducted experiments on a publicly available text-related dataset, as detailed in Table~\ref{tab_public}. Our proposed methods consistently outperformed other specialized medical LLMs and variants of GPT (such as ChatGPT and GPT-4) in both clinical report summarization and \add{radiotherapy strategy} suggestion tasks. Notably, our RO-LMM-P++ method demonstrated substantial performance enhancements, with improvements of up to 0.55 and 0.31 over other medical LLMs and GPT variants, respectively, demonstrating our model's robustness to error accumulation during sequential generations.

\section{Ablation Study}

\subsection{Analysis on Strategy of Separate Expert}
{To validate the rationale behind adopting separate strategies for each textual task, we conduct experiments by preparing another 
Default model that unifies summarization and plan suggestion, and show the results in Table~\ref{tab_text_analysis}. We observe that the unified model's performance is significantly inferior to our distinct training strategy. This implies that summarization task, which provides information within the input, is significantly different from suggesting \add{radiotherapy strategy} that does not exist within the input, indicating the challenging nature of the task. These results suggest that the knowledge required for each task necessitates task-aware training strategies.}\\

\begin{table}[!t]
    \caption{Ablation study on adopting separate expertise for each textual task against unified strategy.} 
    \centering
    \resizebox{0.85\linewidth}{!}{
    \begin{tabular}{ccccccc}
          \toprule
       \multirow{2}{*}{\shortstack[c]{\\ \bf{Training Strategy}  }}& \multicolumn{3}{c}{\bf{Internal }} & \multicolumn{3}{c}{\bf{External }} \\
        \cmidrule(l){2-4} \cmidrule(l){5-7} 
        & R-1 $\uparrow$ & R-2 $\uparrow$ & R-L $\uparrow$  & R-1 $\uparrow$ & R-2 $\uparrow$ & R-L $\uparrow$  \\
         \cmidrule(l){1-1} \cmidrule(l){2-4} \cmidrule(l){5-7} 
         \multicolumn{5}{l}{\bf{Clinical Note Summary}}\\
         \cmidrule(l){1-1} \cmidrule(l){2-4} \cmidrule(l){5-7}
          Unified strategy & 0.516 & 0.357 & 0.515 & 0.327 & 0.215 & 0.326\\
          Distinct strategy & \bf\add{0.783} & \bf\add{0.605} & \bf\add{0.783} & \bf\add{0.788} & \bf\add{0.607} & \bf\add{0.788} \\
         \cmidrule(l){1-1} \cmidrule(l){2-4} \cmidrule(l){5-7}
         \multicolumn{5}{l}{\bf{\add{Radiotherapy Strategy} Suggestion}}\\
         \cmidrule(l){1-1} \cmidrule(l){2-4} \cmidrule(l){5-7}
          Unified strategy & 0.510 & 0.265 & 0.502 & 0.409 & 0.197 & 0.406 \\
          Distinct strategy  & \bf\add{0.655} & \bf\add{0.500} & \bf\add{0.655} & \bf\add{0.615} & \bf\add{0.459} & \bf\add{0.615} \\
    \bottomrule 
    \end{tabular}
   }
    \label{tab_text_analysis}

\end{table}

\begin{table}[!t]
    \caption{Ablation study on CESEG for target segmentation performance with input text variation.} 
    \centering
     \resizebox{1\linewidth}{!}{
    \begin{tabular}{ccccc}
          \toprule
       \multirow{2}{*}{\shortstack[c]{\\ \bf{CESEG}  }}
        & \multirow{2}{*}{\shortstack[c]{\\\bf{Input Text}}} & \multicolumn{3}{c}{\bf{Target Segmentation}} \\
        \cmidrule(l){3-5} 
        & & Dice $\uparrow$ (CI) & \add{surfaceDice} $\uparrow$ (CI) & IoU $\uparrow$ (CI) \\
         \cmidrule(l){1-1} \cmidrule(l){2-2} \cmidrule(l){3-5} 
          \multirow{3}{*}{\xmark} & Ground Truth &  \add{{0.843} {(0.823-0.863)}} & \add{{0.559} {(0.527-0.591)}} & \add{{0.738} {(0.711-0.764)}}   \\
          & Generated & \add{{0.818} {(0.795-0.841)}} & \add{{0.524} {(0.493-0.555)}} & \add{{0.705} {(0.676-0.736)}} \\
          \cmidrule(l){2-2} \cmidrule(l){3-5}  
          & Difference $\downarrow$ & $0.025$ & $0.035$ & $0.033$  \\
         \midrule
          \multirow{3}{*}{\cmark} & Ground Truth  & \add{{0.842} {(0.819-0.861)}} & \add{{0.558} {(0.524-0.589)}} & \add{{0.736} {(0.708-0.761)}} \\
          & Generated & \add{{0.838} {(0.818-0.858)}} & \add{{0.555} {(0.523-0.583)}} & \add{{0.732} {(0.706-0.757)}} \\
          \cmidrule(l){2-2} \cmidrule(l){3-5} 
          & Difference $\downarrow$ & \bf 0.004 & \bf0.003 & \bf0.004 \\
    \bottomrule 
    \end{tabular}
   }
    \label{tab_seg_analysis}
\end{table}

\subsection{Analysis on Consistency Regularization}
{We further analyze the effect of the proposed consistency module on different input text types. 
As mentioned, the segmentation model is trained on clean ground truth data. However, during inference, it relies on potentially noisy data generated by the LMM. This inherent discrepancy between training and inference data can lead to a performance gap.}
As indicated as difference in Table~\ref{tab_seg_analysis}, the model's performance without our proposed consistency module significantly degrades for the generated data compared to the ground truth data for target segmentation tasks. 
{In contrast, our proposed CESEG module demonstrably improves the robustness of the model when using LMM-generated data. We observed a performance difference of less than 1$\%$ between the model's dice score and IoU on generated data and ground truth data. This result suggests that CESEG effectively mitigates performance degradation caused by noisy LMM-generated outputs.}

\subsection{\add{Analysis of Noise type and Intensity ($\alpha$)}} 
\add{Since our proposed method involves injecting noise into embeddings, we conduct a study to analyze the effects of both noise type and intensity under various conditions, as shown in Table~\ref{tab_llm_analysis}. First, regarding noise type, we observe that the method performed best with uniform noise compared to Gaussian noise, leading us to adopt uniform noise as the default configuration. We then examine noise intensity by varying the parameter $\alpha$ from 5 to 15. Lower values of $\alpha$ result in improved performance on both internal and external datasets, while excessively high values caused a noticeable drop in performance on the external dataset.
Building on the main motivation behind NEFTune \cite{jain2024neftune}, we hypothesize that introducing noise to the embedding during training reduces overfitting to the specific characteristics of the instruction fine-tuning dataset, such as formatting details and exact wording. This allows the model to generate responses that incorporate broader knowledge rather than simply replicating the instruction distribution. However, increasing the noise intensity excessively can cause a mismatch between the original data distribution and the test set distribution, underscoring the importance of selecting an optimal noise intensity.
Based on these empirical results, we selected $\alpha = 5$ as the baseline.}

\subsection{\add{Analysis of Finetuning Approach and Model Backbone}} 
\add{In our work, we adopt the full-finetuning approach for instruction finetuning of LLMs. To validate this choice, we conduct experiments comparing full-finetuning to parameter-efficient methods, such as LoRA as indicated in Table~\ref{tab_llm_analysis}. The results showed that using LoRA led to a significant drop in performance on both internal and external datasets. Given the specialized nature of our scope, focused on medical professionals, parameter-efficient finetuning proved insufficient for capturing the target domain knowledge. These findings indicate that full-finetuning is essential for adapting LLMs to the medical domain, particularly in the context of radiation oncology for breast cancer.}

\add{We conduct experiments using the larger 13B LLaMA-2 backbone version and observe a similar performance trend compared to the 7B model as shown in Table~\ref{tab_llm_analysis}. Despite the larger backbone size, our proposed CEFT method outperforms the other baselines, demonstrating scalability. Taking into account both cost-effectiveness and performance, we adopt the 7B model as the default configuration.}

\begin{table}[!t]
    \caption{\add{Component analysis of our proposed method on \add{radiotherapy strategy} suggestion.} } 
    \centering 
    \resizebox{1\linewidth}{!}{
    \begin{tabular}{ccccccccc}   
      \toprule
      \multirow{2}{*}{\shortstack[c]{\bf{Component}}} & \multirow{2}{*}{\shortstack[c]{\bf{Condition}}} & \multirow{2}{*}{\bf{Ours}} & \multicolumn{3}{c}{\shortstack[c]{\bf{Internal Validation }}}  & \multicolumn{3}{c}{\shortstack[c]{\bf{External Validation }}}    \\ 
        \cmidrule(l){4-6} \cmidrule(l){7-9}
        & & &  R-1 $\uparrow$ & R-2 $\uparrow$ & R-L $\uparrow$  &  R-1 $\uparrow$ & R-2  $\uparrow$ & R-L  $\uparrow$  \\
         \cmidrule(l){1-1} \cmidrule(l){2-2} \cmidrule(l){3-3} \cmidrule(l){4-6} \cmidrule(l){7-9}
         \multirow{2}{*}{\shortstack[]{\\{Noise type}}} & \textbf{Uniform 5}  & $\checkmark$ &\bf 0.655 &\bf 0.500 & \bf 0.655 & \bf 0.615 & \bf 0.459  & \bf 0.615 \\
         & Gaussian 5 &  & 0.625 & 0.449 & 0.629 & 0.588& 0.425 & 0.586  \\
         \cmidrule(l){1-1} \cmidrule(l){2-2} \cmidrule(l){3-3} \cmidrule(l){4-6} \cmidrule(l){7-9}
         \multirow{3}{*}{\shortstack[]{\\{Noise Intensity ($\alpha$)}}} & \bf{5}  & $\checkmark$ &\bf 0.655 &\bf 0.500 & \bf 0.655 & \bf 0.615 & \bf 0.459  & \bf 0.615 \\
         & 10 &  & 0.648 & 0.488 & 0.647 & 0.597 & 0.435 & 0.598  \\
         & 15 & & 0.642 & 0.481 & 0.642 & 0.576 & 0.415 & 0.572 \\
         \cmidrule(l){1-1} \cmidrule(l){2-2} \cmidrule(l){3-3} \cmidrule(l){4-6} \cmidrule(l){7-9}
         \multirow{2}{*}{\shortstack[]{\\{Finetuning}}}  & \textbf{Full-finetune}  & $\checkmark$ & \bf 0.655 &\bf 0.500 & \bf 0.655 & \bf 0.615 & \bf 0.459  & \bf 0.615 \\
         & LoRA & & 0.384 & 0.352 & 0.373 & 0.325 & 0.301 & 0.323 \\         
         \cmidrule(l){1-1} \cmidrule(l){2-2} \cmidrule(l){3-3} \cmidrule(l){4-6} \cmidrule(l){7-9}
         \multirow{3}{*}{\shortstack[]{\\{Backbone}}}
         & LLaMA-13B  & & 0.645 & 0.478 & 0.640 & 0.597 & 0.436 & 0.594  \\
         & +NEFT & & 0.658 & 0.500 & 0.653 & 0.601& 0.438 & 0.597 \\
         & \bf{+CEFT} & $\checkmark$ & \bf 0.662 & \bf 0.502 & \bf 0.661 & \bf 0.613 & \bf 0.441  & \bf 0.603 \\
    \bottomrule 
    \end{tabular}
    }
     \label{tab_llm_analysis}
\end{table}

\begin{table}[!t]
    \caption{Inference computational complexity.} 
    \centering
    \resizebox{1\linewidth}{!}{
    \begin{tabular}{lccccc}
          \toprule
         & \multicolumn{4}{c}{\shortstack[c]{\bf{External Validation (N=81) }}} \\
         \cmidrule(lr){2-4} 
            & \bf{RO-LMM-S} & \bf{RO-LMM-P++}  & \bf{RO-LMM-SEG++} & \multicolumn{2}{c}{\bf{Analysis}} \\    
        \cmidrule(l){1-1} \cmidrule(l){2-2}  \cmidrule(l){3-3}  \cmidrule(l){4-4} \cmidrule(l){5-6}  \cmidrule(lr){6-6}
         Memory (GB) & 13.68  & 13.57 & 17.89 & \bf{Mean }&14.93  \\
         Duration (sec/patient)  &  6.89 $\pm$ 1.68 & 1.75 $\pm$ 0.51 & 1.54 $\pm$ 0.72 & \bf{Total} & 10.19 \\
    \bottomrule 

    \end{tabular}
    }
    \label{tile}
\end{table}

\section{Discussion}
While our study provides valuable insights into the application of radiation oncology, several limitations must be acknowledged. Firstly, the dataset is confined to patients with initial breast cancer diagnoses. Expanding to diverse patient scenarios and other cancer types is crucial for broader applicability. Secondly, radiotherapy for breast cancer in real clinical setting often requires techniques like sequential boost or simultaneous integrated boost (SIB) to deliver higher radiation doses around the tumor bed. However, accurate tumor bed delineation requires integrating information from multiple imaging modalities, which simple plan-guided segmentation alone cannot achieve. Therefore, we did not include tumor bed delineation for boost techniques in this study to simplify the problem. Future research should focus on developing a framework capable of delineating these boost volumes to aid in clinical treatment. 
 \add{
Thirdly, the inclusion of non-standard dose prescriptions reflecting country- or institution-specific practices in the training data can limit the generalizability of the results and cause confusion for non-experts. To ensure safer and more broadly applicable clinical implementation, it is crucial to train models using standardized, evidence-based prescriptions.}
Finally, our model did not utilize all clinical information from the actual EMR for decision-making due to token length limitations. We only used MRI, US, and pathology results, considered most critical for initial diagnosis and treatment planning. Overcoming token length constraints to integrate all EMR data will be essential for future studies.

{Despite limitations, our RO-LMM demonstrates the potential to efficiently handle a broad range of clinical workflows using a single NVIDIA A6000 GPU within a 24GB memory, completing tasks in approximately 10 seconds (see Table~\ref{tile}). This capability facilitates local implementation even with limited GPU resources and holds the promise of reducing the workload of radiation oncologists via the automation of clinical report-based plan suggestion for delineating radiation target volumes. }
Furthermore, our method demonstrates its effectiveness by surpassing closed-source LLMs on both a multi-center validation set and a publicly available dataset. These results suggest that our locally tuned RO-LMM-P++ holds significant potential to replace closed-source LLMs while maintaining data privacy and compliance with local data exposure regulations.

\section{Conclusion}
In this work, we introduce RO-LMM, a multi-purpose, comprehensive foundation model tailored for radiation oncology. Addressing limitations in current medical AI models confined to specific tasks, RO-LMM demonstrates proficiency in diverse tasks encompassing overall workflow of radiation oncology: clinical report summarization, \add{radiotherapy strategy} suggestion, and plan-guided 3D target volume segmentation.
Another key contribution of this work is the introduction of consistency technique into both text and segmentation task. Results from multi-center cohort datasets confirm RO-LMM's promising performance and noteworthy generalization capabilities across diverse tasks. These findings mark a significant stride toward developing a versatile AI model, hinting at the potential for a multi-purpose medical AI model in radiation oncology.

\section*{Acknowledgments}
This research was supported by the National Research Foundation of Korea (NRF) grant funded by the Korea government (MSIT) (\#2022R1A2C2008623), and also by the Ministry of Education (\#RS-2023-00242164). This work was also supported by the Korea Health Industry Development Institute (KHIDI) grant funded by the Ministry of Health \& Welfare, Republic of Korea (\#HI23C0730000023). This work was also supported by the National Research Foundation of Korea (NRF) (\#RS-2023-00262527) and the Korea Medical Device Development Fund grant funded by the Korea government (the Ministry of Science and ICT, the Ministry of Trade, Industry and Energy, the Ministry of Health \& Welfare, the Ministry of Food and Drug Safety) (Project Number: 1711137899, KMDF\_PR\_20200901\_0015).

\appendix


\renewcommand{\thefigure}{A\arabic{figure}}
\renewcommand{\thetable}{A\arabic{table}}
\setcounter{figure}{0}
\setcounter{table}{0}

\section{Details of Training and \add{Validation} Dataset} 
\label{sec:appn_dataset}

For data preparation, we gathered internal training and validation data from patients treated at Yonsei Cancer Center between January 2009 and December 2022.

For external validation, data was collected from patients at Yongin Severance Hospital from January 2018 to December 2022, ensuring no overlap with the training set.
For pre-processing, all the chest CT scans and target volumes were re-sampled with identical voxel spacing. 
The Hounsfield unit (HU) of the CT scans were truncated between -1,000 and 1,000, and linearly normalized to the range of 0 to 1.0.
During training, 3D patches was randomly cropped, whereas, during inference, the entire 3D CT scan was tested using sliding windows of the same spatial dimension as the training patches. 
All the patient information and any potential personally identifiable information from used datasets are de-identifiied, and clinical data usage is approved by the Institutional Review Board (IRB).

\section{Details of Open-sourced Dataset} 
\label{sec:open_dataset}

To rigorously evaluate our framework across different hospital settings, we created synthetic patient cases modeled after Gachon University Gil Hospital. These cases covered various breast cancer types (left, right, bilateral), stages (Tis-T3, N0-N3), and included scenarios with breast-conserving surgery, both with and without neoadjuvant chemotherapy, and diverse molecular subtypes. GPT-4 was used to generate clinical notes, preoperative ultrasound, MRI reports, and surgical pathology findings tailored to Gil Hospital’s style. From 50 generated cases, we selected the 20 most realistic ones with the help of a board-certified radiation oncologist. This curated dataset, free of real patient data, is available as an open-source resource at https://github.com/tvseg/ro-lmm.

\section{Detailed Prompt and Rubric for Evaluation } \label{sec:appn_prompt}

 We design prompts for evaluating generated \add{report summarizations and} \add{radiotherapy strategies} using a reference-guided scoring rubric curated by clinical experts, as detailed in Table  \add{\ref{tab:prompt_summary} and ~\ref{tab:plan_example}, respectively}. \add{For the report summarizations, there are four evaluation criteria, with the first criterion having a weight of 2, making the total possible score 5. For the \add{radiotherapy strategy} suggestions, there are five criteria, each contributing a score of 1, resulting in a total score of 5.}
Our approach builds on previous GPT-based evaluation methods and integrates clinical experts' knowledge into the prompts. Using these tailored prompts, we assess the performance of our RO-LMM model and all Default methods with GPT-4.0.

\begin{table}[!hbt]
\caption{The proposed expertise-based rubrics for assessing the performance of clinical report summarization.} 
 \centering
 \resizebox{1\linewidth}{!}{
\begin{tabular}{p{3.5cm}p{7.5cm}}
\toprule
\multicolumn{2}{l}{Based on rubrics, score each clinical report summarization.}\\ \\
Rubric 1: Relevance & Assessment of whether the necessary information from the Ground Truth (GT) is included, and identification of any extraneous information \\
&Good (2): The information from the GT is appropriately included.\\
&Partially correct (1): Some relevant information is included, but there are omissions or irrelevant details.\\
&Mostly irrelevant (0): The majority of the information is unnecessary or irrelevant.\\
Rubric 2: Formatting &  Evaluation of whether the necessary information is presented in the correct format and location:\\
&Incorrectly placed (0): Information is not presented in the appropriate format or location.\\
&Correctly placed (1): The necessary information is accurately positioned in the expected format.\\
Rubric 3: Length & Assessment of whether the writing is excessively verbose:\\
&Too verbose (0): The writing is overly lengthy and lacks conciseness.\\
&Sufficiently summarized (1): The content is effectively summarized without unnecessary verbosity.\\
Rubric 4: Consistency &Identification of conflicting information between sources:\\
&Present (0): Conflicting pieces of information are observed.\\
&Absent (1): There are no conflicting pieces of information between sources. \\
\bottomrule
\end{tabular}
}
\label{tab:prompt_summary}
\end{table}%

\begin{table}[!hbt]
\caption{Score rubrics for \add{radiotherapy strategy} suggestion.} 

\label{table_prompt}
 \centering
 \resizebox{1\linewidth}{!}{
\begin{tabular}{p{3.3cm}p{7.5cm}}
\toprule

[Rubric (R)] \\
R1. Laterality. &Whether orientation (Rt./Lt.) is correct\\
R2. Aim of Surgery. &Whether aim of RT (definitive RT = RT after breast conservation surgery or partial mastectomy. In this scenario, breast should be irradiated. / postop RT = RT after total or radical mastectomy. In this scenario, chest wall should be irradiated) is correct.\\
R3. Treatment Scope. &Whether detailed extent of radiation field (inclusion or exclusion of WB, CW, AXL, SCL, IMN) and  detailed level of inclusion (RTOG = in this guideline, upper SCL node is included / ESTRO = in this guideline, upper SCL is not included) is correct.\\
R4. Dose Scheme. &Whether RT scheme (dose in Gy, and fractions in fx) is correct. It also included SIB or sequential boost dose.\\
R5. Hallucination. &Whether unnecessary information is  added or not (If unnecessary \& incorrect information was added, score 0 points, otherwise score 1 point)\\

\multicolumn{2}{l}{[Abbreviations]} \\
RT. &Radiotherapy\\
BCS. &Breast Conserving Surgery\\
WB. &Whole Breast\\
CW. &Chest Wall\\
AXL. &Axilla Lymph Node\\
SCL. &Supraclavicular Lymph Node\\
IMN. &Internal Mammary Lymph node\\
RTOG. & Radiation Therapy Oncology Group \\
ESTRO. &European Socity for Radiotherapy and Oncology\\
SIB. &Simultaneous Integrated Boost technique, same as tumor bed boost (sequential), fast-forward protocol\\ \\

\multicolumn{2}{l}{[Ground Truth Answer Begin]} \\
\multicolumn{2}{l}{$<$Ground Truth Plan$>$} \\ 
\multicolumn{2}{l}{[Ground Truth Answer End]} \\ \\ 

\multicolumn{2}{l}{[Model Responses Begin]} \\ 
\multicolumn{2}{l}{$<$Model Generated Plan$>$} \\ 
\multicolumn{2}{l}{[Model Responses End]} \\ \\  

\multicolumn{2}{l}{[Answer Format]}\\
\multicolumn{2}{l}{Model, Answer for R1, Answer for R2, Answer for R3, Answer for R4, Answer}\\
\multicolumn{2}{l}{for R5, scores ONLY for each Rubric in a tabular format, without explanation:}\\

\bottomrule
\end{tabular}
}
\label{tab:plan_example}
\end{table}%

\bibliographystyle{splncs04}
\bibliography{refs}

\end{document}